\documentclass{article} 
\usepackage{iclr2026_delta,times}

\usepackage{amsmath,amsfonts,bm}

\def\eqref#1{equation~\ref{#1}}

\def\1{\bm{1}}

\DeclareMathAlphabet{\mathsfit}{\encodingdefault}{\sfdefault}{m}{sl}
\SetMathAlphabet{\mathsfit}{bold}{\encodingdefault}{\sfdefault}{bx}{n}

\usepackage{hyperref}
\usepackage{url}
\usepackage{graphicx}
\usepackage{amsmath}
\usepackage{amssymb}
\usepackage{multirow}
\usepackage{tabularx}
\usepackage{float}
\usepackage{xcolor}

\iclrfinalcopy
\title{AnimalBooth: Multimodal Feature Enhancement for Animal Subject Personalization}

\author{Chen Liu \& Haitao Wu \\
College of Intelligence and Computing\\
Tianjin University\\
Tianjin, China \\
\texttt{\{ttjy, wuhaitao\}@tju.edu.cn} \\
\AND
Kafeng Wang \\
Department of Computer Science and Technology\\
Tsinghua University\\
Beijing, China \\
\texttt{wangkafeng@tsinghua.edu.cn} \\
\AND
Weiran Huang\thanks{Corresponding author.} \\
School of Computer Science\\
Shanghai Jiao Tong University\\
Shanghai, China \\
\texttt{weiran.huang@sjtu.edu.cn} 
}

\begin{document}

\maketitle

\begin{abstract}
\looseness=-1
Personalized animal image generation is challenging due to rich appearance cues and large morphological variability. Existing approaches often exhibit feature misalignment across domains, which leads to identity drift. We present \textit{AnimalBooth}, an inference-time tuning-free framework that strengthens identity preservation with an \textit{Animal-Net} and an adaptive attention module, mitigating cross-domain alignment errors. We further introduce a frequency-controlled feature integration module that applies Discrete Cosine Transform filtering in the latent space to guide the diffusion process, enabling a coarse-to-fine progression from global structure to detailed texture. To advance research in this area, we curate \textit{AnimalBench}, a high-resolution dataset for animal personalization. Extensive experiments show that \textit{AnimalBooth} consistently outperforms strong baselines on multiple benchmarks with superior efficiency, improving both identity fidelity and perceptual quality. The code and dataset will be made publicly available in the future.
\end{abstract}

\section{Introduction}
\label{sec:intro}
Personalized multimodal generation is a prominent yet challenging subfield that aims to synthesize images conforming to both textual descriptions (text--image consistency) and the intrinsic characteristics of custom concepts (identity consistency)~\citep{deng2025locref, dai2025diffusefist, PersonalizedICASSP, shen2025imagharmony}. This paradigm shows strong potential across diverse applications ranging from creative artistry to product design~\citep{shen2025imaggarment, shen2025long}. However, accurately capturing the identity of animal subjects, which exhibit complex visual features such as fine-grained fur textures and non-rigid morphologies, remains a significant hurdle~\citep{Anima2ICASSP, shen2024imagpose}.

Animal personalization is uniquely challenging compared to human subject personalization due to several factors: (1) \textbf{Non-rigid deformations}: animals exhibit diverse poses and body configurations that vary dramatically across species; (2) \textbf{Anatomical variations}: skeletal structures and body proportions differ significantly between species (e.g., felines vs. bovines); (3) \textbf{Intricate texture requirements}: fur patterns, scales, and feathers require fine-grained preservation that general-purpose models often fail to maintain. As demonstrated in recent work~\citep{Anima2ICASSP}, general-purpose models like DiT-based architectures frequently suffer from identity drift in these challenging scenarios.

Existing personalization methodologies can be broadly categorized by their feature extraction and optimization strategies. The first category comprises optimization-based methods that achieve high fidelity by fine-tuning model components. A representative example is DreamBooth~\citep{DB}, which fine-tunes the UNet backbone, while Textual Inversion~\citep{gal2022image} instead optimizes word embeddings. This paradigm has been extended to support multiple customized concepts~\citep{kumari2023custom, ma2024subject, wang2025msdiffusionmultisubjectzeroshotimage}, including applications such as virtual dressing and editing~\citep{shen2025imagdressing}. Despite their fidelity, such approaches demand substantial computational resources and often exhibit an inherent trade-off between fidelity and diversity due to overfitting.  
The second category, encoder-based and tuning-free methods, offers a more efficient alternative. For instance, IP-Adapter~\citep{ye2023ipadapter} and its successors~\citep{zhang2023ssrencoder} utilize frozen external encoders (e.g., CLIP) to inject identity features. While efficient, their performance is fundamentally limited by the representational capacity of the encoder, particularly when modeling animals with unique identity cues such as body structures and fur patterns. Consequently, identity distortion and detail loss are frequent. Other directions include retrieval-augmented generation~\citep{chen2022reimagen} or LLM-enhanced frameworks~\citep{zeng2024jedi, pan2023kosmosg, Sun_2024_CVPR}, yet the challenge of inadequate domain-specific representation remains.

\looseness=-1
Motivated by bridging this domain gap, we propose \textbf{AnimalBooth}, a feature-enhanced, \textbf{inference-time tuning-free} personalized generation framework specifically tailored for animals. To systematically address these identified challenges, our framework incorporates several targeted mechanisms: (1) To handle \textit{non-rigid deformations}, we introduce a frequency-controlled feature integration module that leverages low-frequency signals to guide coarse structural consistency; (2) To overcome \textit{encoder capacity} limits and \textit{identity drift}, we design a dedicated Animal-Net with a Q-Former bottleneck to filter noise and capture subject-specific semantic features; (3) To maintain \textit{text-control} while injecting identity, we employ a dual-path adaptive attention mechanism.

Furthermore, existing datasets such as AFHQ~\citep{AFHQ} or MS-COCO are primarily designed for classification, translation, or general captioning, and lack the high-resolution subject masks and fine-grained captions necessary for precise animal personalization. To address this, we construct \textbf{AnimalBench}, a curated high-definition dataset for animal personalization.
Our contributions are threefold:  
(1) We propose a dedicated branch architecture comprising a lightweight Animal-Net and a novel dual-path adaptive attention mechanism, effectively mitigating identity distortion while preserving generative capacity.  
(2) We design a frequency-controlled feature integration module leveraging DCT filtering in latent space to enhance attribute manipulation and texture fidelity.  
(3) We establish and release AnimalBench, a high-definition dataset for animal personalization, on which our method achieves state-of-the-art performance while being significantly more efficient than large-scale DiT-based models.

\begin{figure}[t]
\begin{center}
\includegraphics[width=1.0\linewidth]{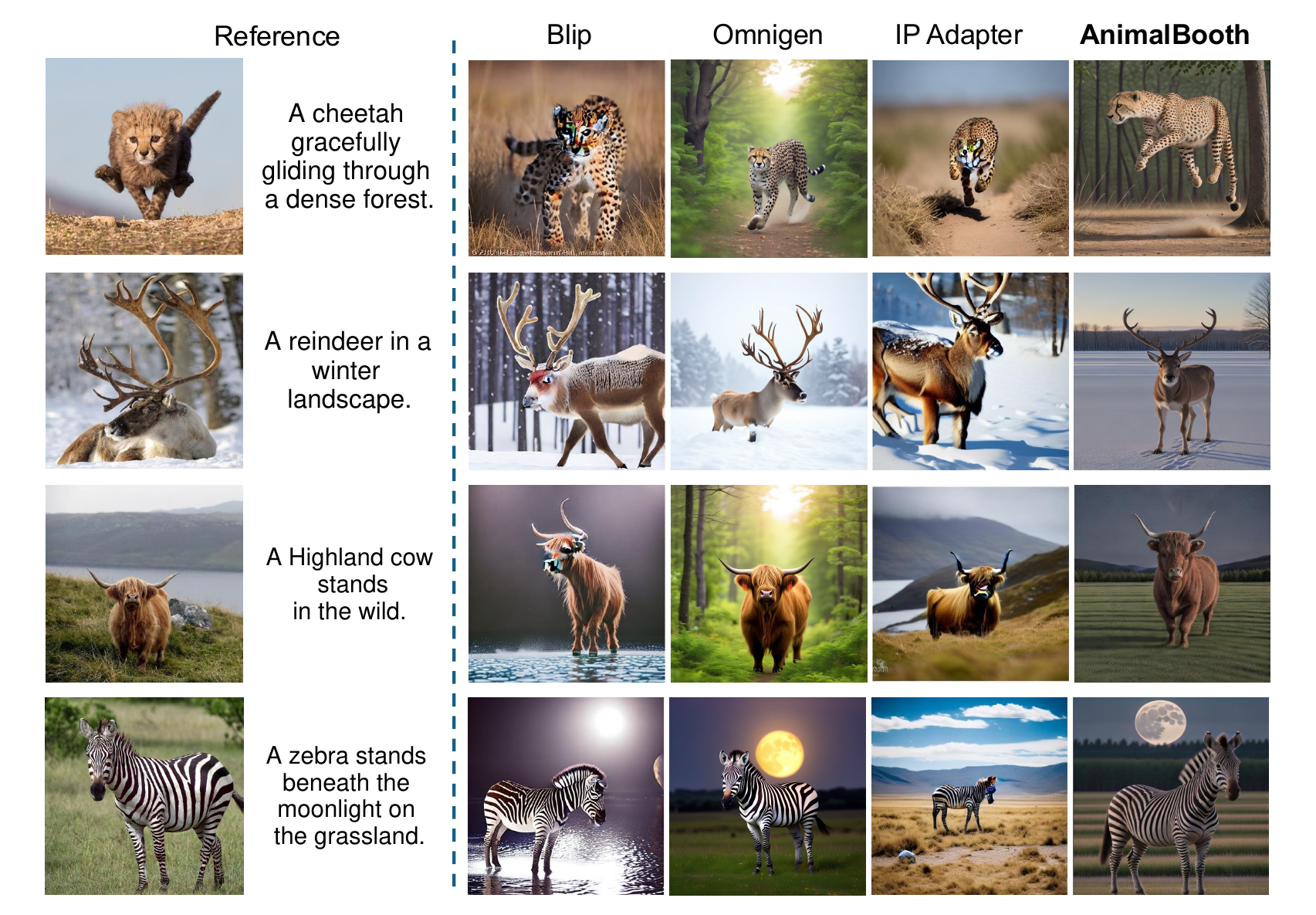}
\end{center}
\vspace{-3mm}
  \caption{Qualitative comparison with SOTA methods.} 
  \label{fig:qualitative}
\end{figure}

\begin{figure*}[t!]
\begin{center}
  \includegraphics[width=0.95\linewidth]{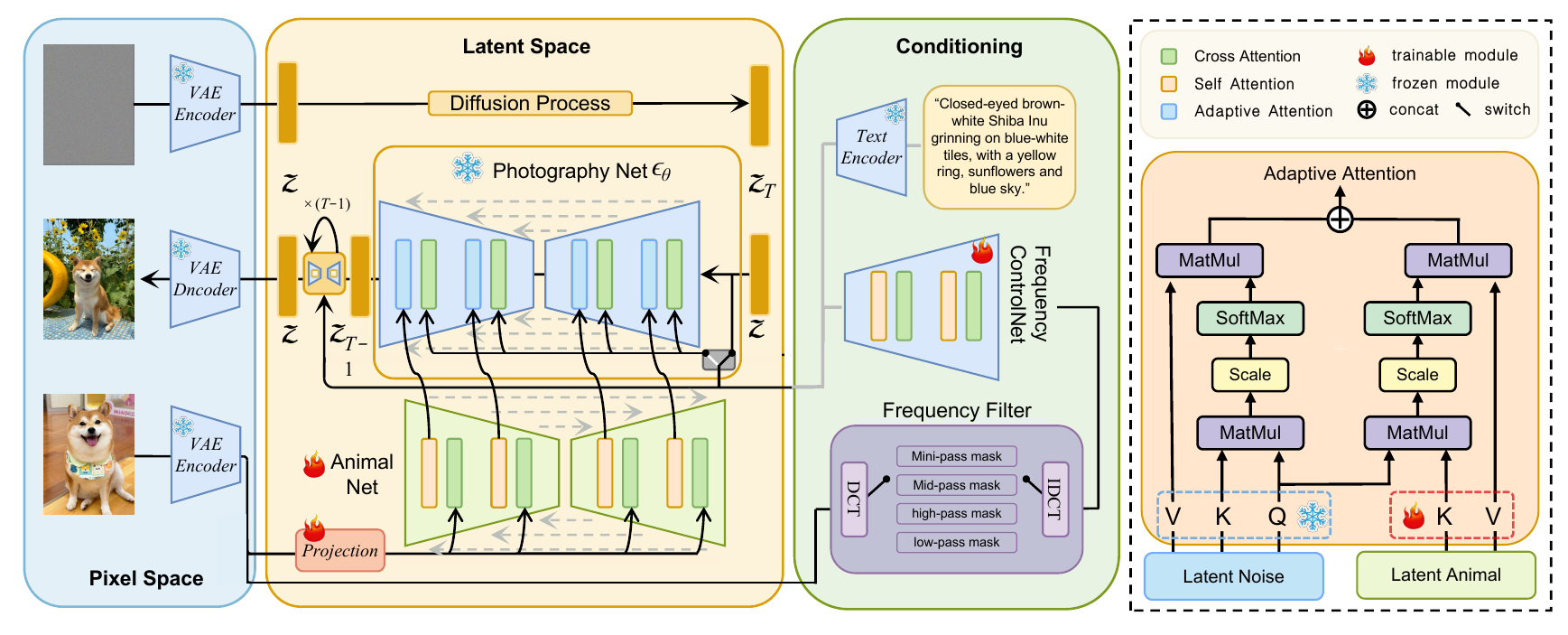}
\end{center}
  \caption{AnimalBooth primarily consists of a trainable Animal-Net and a frozen Photography-Net. The Animal-Net incorporates an Adaptive Attention module for efficient identity feature injection and a Frequency-Controlled module (based on ControlNet~\citep{zhang2023controlnet}) for enhanced control over visual attributes like structure and texture, while the Photography-Net integrates these features with text prompts within the latent space.}
\vspace{-2mm}
  \label{fig:structure}
\end{figure*}

\begin{figure}[t]
\begin{center}
\includegraphics[width=1.0\linewidth]{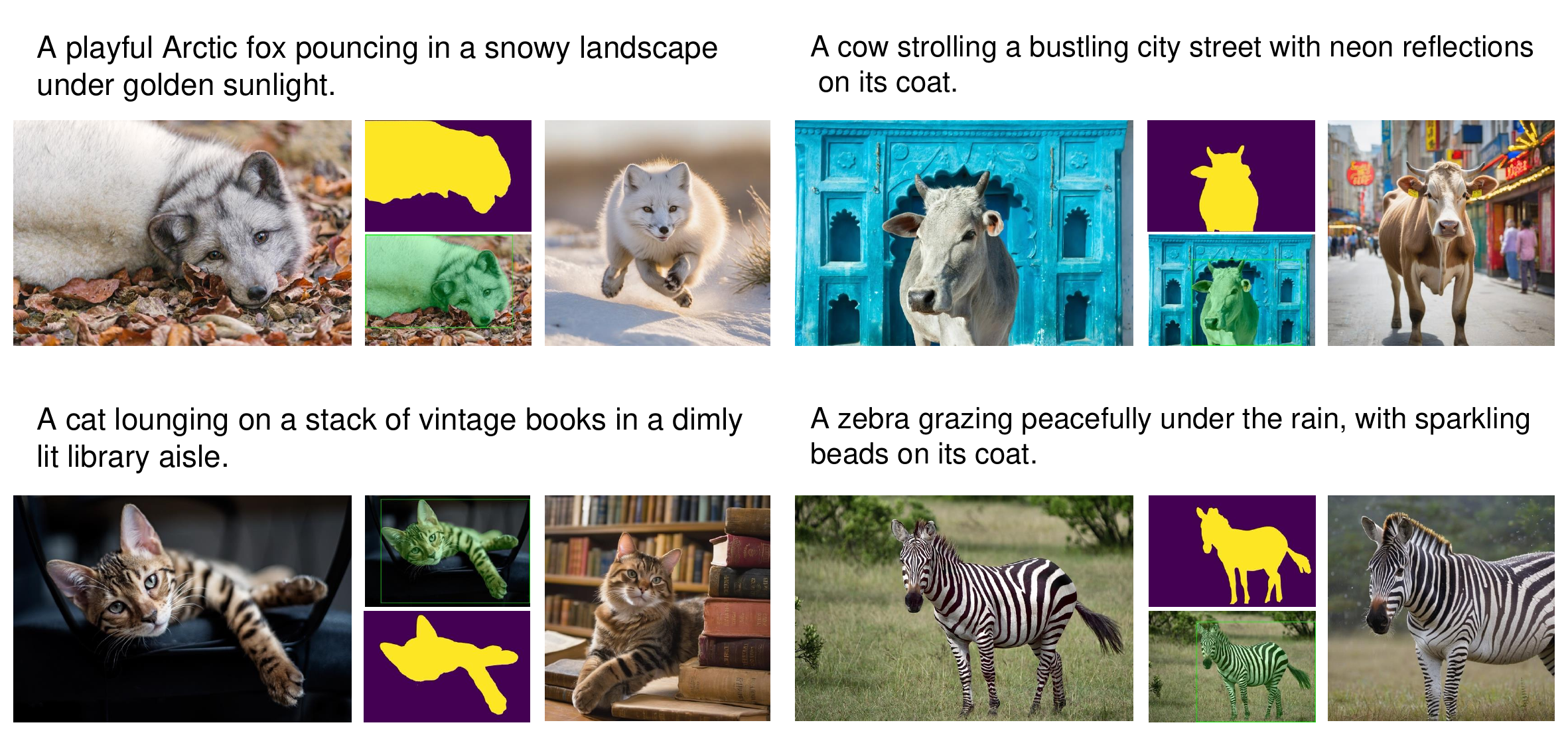}
\end{center}
\vspace{-3mm}
  \caption{Examples of AnimalBench dataset. Each instance provides five components: (a) a high-quality source image, (b) a detailed caption describing the scene and subject, (c) a precise pixel-level semantic segmentation mask of the primary subject, (d) the resulting masked subject for visual verification, and (e) the corresponding paired image.}
\vspace{-3mm}
  \label{fig:dataset}
\end{figure}

\section{Methodology}
\label{sec:method}
\subsection{Overall Architecture}
\label{ssec:overall_arch}
As depicted in Fig.~\ref{fig:structure}, AnimalBooth effectively integrates a trainable Animal-Net with a frozen Photography-Net. The Animal-Net directly refines the complex features and flexible morphologies of animals within the feature space, effectively bypassing the complexities of cross-domain alignment. Through adaptive attention, identity features are efficiently injected. The incorporation of a frequency-controlled module based on ControlNet~\citep{zhang2023controlnet} further enhances fine-grained control over visual attributes such as the structure and textural details of the generated images. The frozen Photography-Net then merges these extracted fine-grained features with text prompts in the latent space.

\subsection{Animal-Net}
In the task of personalized animal image generation, precise extraction of fine-grained animal identity features is paramount for maintaining consistency between generated images and reference animals. To achieve this, we propose a specialized Animal-Net capable of simultaneously capturing both semantic information and high-frequency textural features of animals. Specifically, given an animal reference image $\mathcal{X}_a \in \mathbb{R}^{3 \times H \times W}$, we first transform it into a latent space representation $\mathbf{L}_a \in \mathbb{R}^{4 \times \frac{H}{8} \times \frac{W}{8}}$ using a frozen VAE encoder~\citep{rombach2022high}. We then extract token embeddings from $\mathcal{X}_a$ using a frozen CLIP image encoder~\citep{radford2021learning} and a trainable projection layer. 

\noindent\textbf{Why Q-Former?} We employ Q-Former~\citep{li2023blip} as the projection layer due to its superior ability to act as a semantic bottleneck that effectively filters background noise while capturing intricate animal textures via dynamic attention. Unlike static mapping approaches (e.g., feature concatenation or MLP adapters), Q-Former's learnable query tokens dynamically attend to relevant identity features. As shown in Table~\ref{tab:projection}, Q-Former significantly outperforms simpler alternatives in identity fidelity metrics.

Subsequently, animal features within the Animal-Net interact extensively through a cross-attention mechanism, similar to the interaction between text and image features in the original Text-to-Image (T2I) model~\citep{rombach2022high}. This interaction ensures a deep fusion of semantic and textural features. Finally, the output of the Animal-Net is aligned in parallel with the Photography-Net, injecting these fine-grained animal identity features into the Photography-Net via an adaptive attention module. Notably, the Animal-Net is solely used for encoding reference images; therefore, during the diffusion process, no noise is added to the reference image, and it undergoes only a single forward pass.

\subsection{Adaptive Attention Module}
To enable personalized animal image generation, the Photography-Net must possess both its original generative capabilities and the ability to fuse animal identity features. We freeze the core modules of the Photography-Net to preserve the former, and achieve the latter through an adaptive attention module. The architecture of the Photography-Net in AnimalBooth builds upon SD v1.5~\citep{rombach2022high}, with all self-attention modules replaced by adaptive attention modules. 

\noindent\textbf{Dual-Path Architecture.} Unlike standard single-path cross-attention strategies commonly used in subject-driven generation, our adaptive attention module employs a novel \textbf{dual-path architecture}: a frozen self-attention path (to preserve the original generative capacity of Stable Diffusion) and a trainable cross-attention path (to inject identity features), both sharing a single Query matrix $\mathbf{Q}$. This shared-query design enables seamless feature fusion while maintaining the pre-trained model's semantic understanding. As shown in Fig.~\ref{fig:structure}, an adaptive attention module consists of a frozen self-attention module and a learnable cross-attention module. Its self-attention weights are initialized from SD v1.5 to retain generative capacity. Given the query features $\mathbf{Z}_n$ from the Photography-Net and the animal identity features $\mathbf{F}_a$ from the Animal-Net, the output $\mathbf{O}_h$ of the adaptive attention module is defined as:
\begin{equation}
\mathbf{O}_{h} =
\underbrace{\operatorname{Softmax}\left(\frac{\mathbf{Q} \mathbf{K}^{\top}}{\sqrt{d}}\right) \mathbf{V}}_{\text{Frozen Self-attention}} +
\lambda\underbrace{\operatorname{Softmax}\left(\frac{\mathbf{Q}\left(\mathbf{K}_{ID}\right)^{\top}}{\sqrt{d}}\right) \mathbf{V}_{ID}}_{\text{Trainable Cross-attention}} ~,
\label{eq:adaptive_attention}
\end{equation}
where $\lambda \in [0, 1]$ is a dynamic coefficient that controls the strength of the animal identity feature condition, enabling flexible balance between identity preservation and creative generation. $\mathbf{Q}$, $\mathbf{K}$, $\mathbf{V}$ are derived from $\mathbf{Z}_n$, while $\mathbf{K}_{ID} = \mathbf{F}_a \mathbf{W}_{k_{ID}}$ and $\mathbf{V}_{ID} = \mathbf{F}_a \mathbf{W}_{v_{ID}}$ are derived from $\mathbf{F}_a$. The self-attention part is frozen, while the cross-attention part (i.e., $\mathbf{W}_{k_{ID}}$ and $\mathbf{W}_{v_{ID}}$) is trainable. This design effectively injects animal identity features while preserving the original T2I model's~\citep{rombach2022high} generative capabilities.

\subsection{Frequency-Controlled Feature Integration Module}
This module guides the diffusion process in the latent space through Discrete Cosine Transform (DCT) filtering, implemented via ControlNet~\citep{zhang2023controlnet} (Fig.~\ref{fig:structure}). First, a channel-wise 2D DCT is applied to the source domain latent features $\mathbf{L}_0$ (obtained from the VAE encoder) to acquire the frequency domain representation $\mathbf{F}_{DCT}$:
\begin{equation}
\begin{aligned}
    \mathbf{F}_{DCT,u,v}^{(n)}&=\frac{2}{\sqrt{hw}}m(u)m(v)\sum\nolimits_{i=0}^{h-1}\sum\nolimits_{j=0}^{w-1}[(\mathbf{L}_{0}^{(n)})_{i,j} \\
    &\cos\left(\frac{(2i+1)u\pi}{2h}\right)\cos\left(\frac{(2j+1)v\pi}{2w}\right)],
\end{aligned}
\end{equation}

where $m(0)=\frac{1}{\sqrt{2}}$ and $m(\gamma)=1$ for ($\gamma>0$).

Different frequency bands of the DCT spectrum encode distinct visual attributes: low-frequency components capture global structure and identity-related morphology, while high-frequency components encode fine textures and edges. We design four types of DCT filters (masks) for mini-pass, low-pass, mid-pass, and high-pass filtering to manipulate visual properties from coarse structures to fine textures.
\begin{equation}
    \begin{cases}
        Mask_{mini}(u,v)=1\ \ if\ \ u+v\leq10\ \ else\ \ 0, \\
        Mask_{low}(u,v)=1\ \ if\ \ u+v\leq20\ \ else\ \ 0, \\
        Mask_{mid}(u,v)=1\ \ if\ \ 20<u+v\leq40\ \ else\ \ 0, \\
        Mask_{high}(u, v)=1\ \ if\ \ u+v\geq50\ \ else\ \ 0.
    \end{cases}
\end{equation}
These filters are multiplied by $\mathbf{F}_{DCT}$ to extract features in specific frequency bands, denoted as $\mathbf{F}_{filtered}=\mathbf{F}_{DCT}\times Mask_{*}$. Finally, a 2D Inverse DCT (IDCT) is applied to convert $\mathbf{F}_{filtered}$ back to the spatial domain, yielding the control signal $\mathbf{C}_{freq}$:
\begin{equation}
\begin{aligned}
    \mathbf{C}_{\mathrm{freq},\,i,j}^{(n)}
    &= \frac{2}{\sqrt{hw}}
       \sum_{u=0}^{h-1}\sum_{v=0}^{w-1}
       m(u)\,m(v)\,
       \bigl(\mathbf{F}_{\mathrm{filtered}}^{(n)}\bigr)_{u,v}\\
    &\quad \times
       \cos\!\left(\frac{(2i+1)u\pi}{2h}\right)
       \cos\!\left(\frac{(2j+1)v\pi}{2w}\right).
\end{aligned}
\end{equation}
The $\mathbf{C}_{freq}$ signals obtained from mini-pass, low-pass, mid-pass, and high-pass filtering, respectively, control the texture, texture and structure, layout, and contour consistency between the generated and reference images. As our experiments demonstrate (Table~\ref{tab:ablation}), low-pass filtering achieves optimal identity preservation by retaining the structure and morphology information critical for animal recognition.

\subsection{Training and Objectives}
AnimalBooth training is bifurcated into two complementary stages that follow a coarse-to-fine progression:

\noindent\textbf{Stage 1: Identity Learning.} The Animal-Net and Projection modules are trained to capture global subject identity. The objective function aims to minimize the Mean Squared Error between the predicted noise and the ground truth noise, while incorporating both text conditions $\mathbf{C}_t$ and animal identity features $\mathbf{C}_a$:
\begin{equation}
 L_{stage1} = \mathbb{E}_{\mathbf{z}_t, \epsilon \sim \mathcal{N}(\mathbf{0}, \mathbf{I}), \mathbf{C}_t, \mathbf{C}_a, t}\left\|\epsilon_\theta\left(\mathbf{z}_t, \mathbf{C}_t, \mathbf{C}_a, t\right)-\epsilon_t\right\|^2.
\end{equation}

\noindent\textbf{Stage 2: Texture Enhancement.} The frequency-controlled ControlNet module is trained to enhance fine textures and structural details. This decoupled training strategy avoids convergence difficulties that arise from jointly optimizing identity and texture objectives:
\begin{equation}
    L_{stage2} = \mathbb{E}_{\mathbf{z}_0, t, \mathbf{C}_t, \mathbf{C}_{freq}, \epsilon \sim \mathcal{N}(\mathbf{0}, \mathbf{I})}\left\|\epsilon - \epsilon_\theta(\mathbf{z}_{t}, t, \mathbf{C}_t, \mathbf{C}_{freq})\right\|_{2}^{2},
\end{equation}
where $\mathbf{z}_t$ represents the noisy latent features at time step $t$, $\mathbf{C}_t$ is the text condition, and $\mathbf{C}_{freq}$ is computed by the frequency-controlled feature integration module based on the source domain latent features $\mathbf{L}_0$. The parameters of ControlNet are trained using four distinct DCT filters ($Mask_{mini}, Mask_{low}, Mask_{mid}, Mask_{high}$).

During inference, classifier-free guidance~\citep{ho2021classifier} is employed to balance conditional and unconditional generation:
\begin{equation}
\begin{split}
\hat{\epsilon}_{\theta}(\mathbf{x}_t, \mathbf{C}_t, \mathbf{C}_a, \mathbf{C}_{freq}, t) &= w \epsilon_{\theta}(\mathbf{x}_t, \mathbf{C}_t, \mathbf{C}_a, \mathbf{C}_{freq}, t) + (1 - w) \epsilon_{\theta}(\mathbf{x}_t, t),
\end{split}
\end{equation}
where $w$ is the guidance scale. We empirically set $w=7.5$ in our experiments (see Table~\ref{tab:guidance} for ablation).

\begin{figure*}[b!]
\begin{center}
\includegraphics[width=0.9\linewidth]{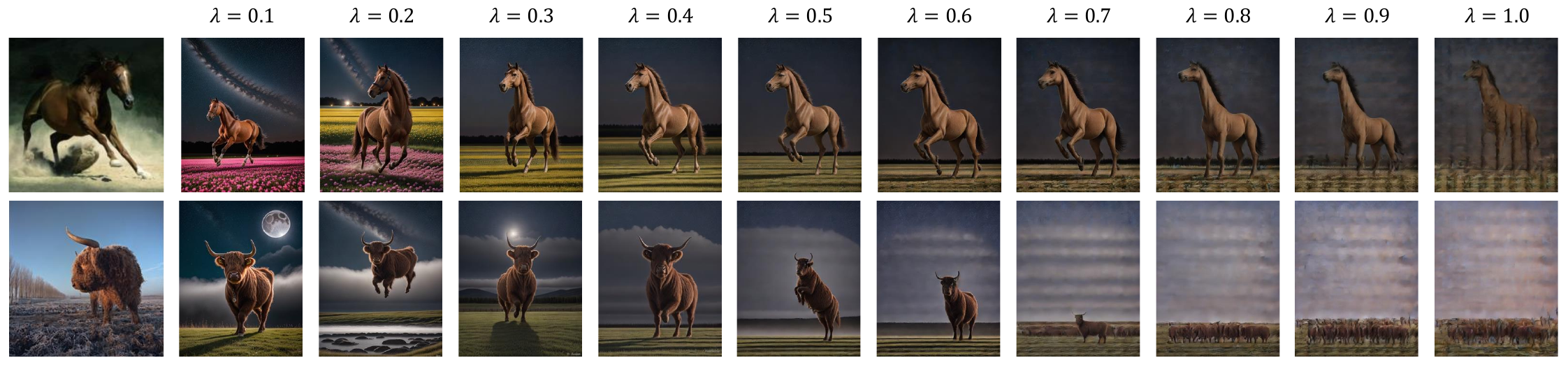}
\end{center}
\vspace{-2mm}
    \caption{Example results with different animal image strength $\lambda$.}
    \label{fig:image_scale}
\end{figure*}

\section{Experiments}
\label{sec:exp}

\subsection{Experimental Setup}
\label{ssec:setup}

We utilize Stable Diffusion v1.5~\citep{rombach2022high} as the pre-trained Latent Diffusion Model (LDM) and train it for personalized animal image generation. To comprehensively evaluate the model's capabilities in generating high-definition animal images, we constructed a specialized AnimalBench dataset, comprising 10,958 training images and 1,000 test images. Each entry in the dataset consists of a high-definition, single-subject animal image--text pair, with examples depicted in Fig.~\ref{fig:dataset}. Training was conducted on a server equipped with eight NVIDIA A100-SXM4-80GB GPUs. We employed the AdamW optimizer with an initial learning rate of 1e-5 and a batch size of 4. The model was trained at a resolution of 512$\times$512 pixels. During the inference phase, we used the DDIM sampler~\citep{ho2020denoising} for 50 sampling steps with guidance scale $w=7.5$.

\subsection{Qualitative Results}
\looseness=-1
As shown in Fig.~\ref{fig:qualitative}, methods like BLIP-Diffusion~\citep{BLIP-Diffusion}, Omnigen~\citep{xiao2025omnigen} and IP-Adapter~\citep{ye2023ipadapter} struggle with maintaining fidelity to the original animal's identity. For instance, in the ``cheetah'' example, BLIP-Diffusion introduces noticeable distortions in the cheetah's facial features and fur patterns. Omnigen, while better, still presents a cheetah that looks somewhat less dynamic and natural compared to the reference. IP-Adapter's cheetah appears to have some color inaccuracies and a less refined texture. Similar issues are observable in the other examples. For the ``reindeer,'' these methods often fail to capture the subtle nuances of its fur, antler structure, or the serene expression seen in the reference, sometimes introducing unnatural postures or color shifts. The ``Highland cow'' examples highlight a lack of accurate texture and the distinct shagginess that characterizes the breed. Finally, the ``zebra'' images from these comparative methods often fall short in reproducing the sharp stripe patterns, the reflective quality of its fur under moonlight, or the natural stance, sometimes resulting in blurry details or less vibrant contrasts. AnimalBooth consistently demonstrates superior performance in preserving both coarse structures and fine-grained textures.

\begin{table}[t]
\centering
\begin{minipage}{0.49\textwidth}
\centering
\caption{Ablation on projection layer.}
\label{tab:projection}
\resizebox{\linewidth}{!}{
\begin{tabular}{lcccc}
\multicolumn{1}{c}{\bf Projection}  &\multicolumn{1}{c}{\bf LPIPS $\downarrow$} &\multicolumn{1}{c}{\bf DINO $\uparrow$} &\multicolumn{1}{c}{\bf CLIP-T $\uparrow$} &\multicolumn{1}{c}{\bf CLIP-I $\uparrow$}
\\ \hline \\
Concatenation & 57.41 & 31.05 & 19.85 & 70.58 \\
MLP Adapter & 56.81 & 31.68 & 20.02 & 71.00 \\
Q-Former (Ours) & \textbf{49.08} & \textbf{75.66} & \textbf{20.73} & \textbf{90.00} \\
\end{tabular}
}
\end{minipage}
\hfill
\begin{minipage}{0.49\textwidth}
\centering
\caption{Ablation on guidance scale $w$.}
\label{tab:guidance}
\resizebox{\linewidth}{!}{
\begin{tabular}{ccccc}
\multicolumn{1}{c}{\bf $w$}  &\multicolumn{1}{c}{\bf LPIPS $\downarrow$} &\multicolumn{1}{c}{\bf DINO $\uparrow$} &\multicolumn{1}{c}{\bf CLIP-T $\uparrow$} &\multicolumn{1}{c}{\bf CLIP-I $\uparrow$}
\\ \hline \\
2.0 & 54.03 & 43.95 & 18.92 & 76.52 \\
5.0 & 51.26 & 68.34 & 20.15 & 85.67 \\
7.5 (Ours) & \textbf{49.08} & \textbf{75.66} & \textbf{20.73} & \textbf{90.00} \\
10.0 & 50.12 & 72.45 & 20.58 & 88.34 \\
\end{tabular}
}
\end{minipage}
\end{table}

\begin{table}[t]
\caption{Efficiency comparison on NVIDIA A100.}
\label{tab:efficiency}
\begin{center}
\begin{tabular}{lccc}
\multicolumn{1}{c}{\bf Method}  &\multicolumn{1}{c}{\bf Time (s) $\downarrow$} &\multicolumn{1}{c}{\bf VRAM (GB) $\downarrow$} &\multicolumn{1}{c}{\bf Params (B)}
\\ \hline \\
Omnigen~\citep{xiao2025omnigen} & 140.0 & 10.4 & 3.8 \\
Flux (DiT)~\citep{chang2024fluxfastsoftwarebasedcommunication} & 85.0 & 12.8 & 12.0 \\
AnimalBooth & \textbf{5.5} & \textbf{0.7} & \textbf{1.2} \\
\end{tabular}
\end{center}
\vspace{-0.2cm}
\end{table}

\begin{table}[t]
\caption{Ablation study on frequency configurations.}
\label{tab:ablation}
\begin{center}
\begin{tabular}{lcccc}
\multicolumn{1}{c}{\bf Method}  &\multicolumn{1}{c}{\bf LPIPS $\downarrow$} &\multicolumn{1}{c}{\bf DINO $\uparrow$} &\multicolumn{1}{c}{\bf CLIP-T $\uparrow$} &\multicolumn{1}{c}{\bf CLIP-I $\uparrow$}
\\ \hline \\
w/o Freq Cond & 69.40 & 58.52 & \textbf{21.37} & 78.17 \\
High-Pass & \underline{56.72} & \underline{64.72} & \underline{21.21} & 80.02 \\
Mid-Pass & 61.66 & 66.97 & 20.84 & \underline{82.79} \\
Mini-Pass & 68.21 & 59.97 & 21.20 & 80.89 \\
Low-Pass (Ours) & \textbf{49.08} & \textbf{75.66} & 20.73 & \textbf{90.00} \\
\end{tabular}
\end{center}
\end{table}

\subsection{Quantitative Results}
\looseness=-1
As presented in Table~\ref{tab:quantitative}, AnimalBooth obtained a CLIP-T~\citep{radford2021learning} score of 20.73, which is notably higher than the next best method, BLIP-Diffusion~\citep{BLIP-Diffusion}, at 19.68. This indicates AnimalBooth's superior ability to comprehend and capture textual semantics, generating animal images that are highly consistent with their descriptions. Concurrently, AnimalBooth achieved the highest DINO~\citep{caron2021emerging} score of 75.66, compared to IP-Adapter's~\citep{ye2023ipadapter} 72.88. This highlights AnimalBooth's exceptional capability in capturing image details and realism, effectively preserving the intricate textures and fine structures of animals. Furthermore, AnimalBooth also achieved leading scores of 90.00 and 49.08 for CLIP-I and LPIPS, respectively, comprehensively outperforming other methods.

\begin{table}[t]
\caption{Quantitative comparison with state-of-the-art methods.}
\label{tab:quantitative}
\begin{center}
\begin{tabular}{lcccc}
\multicolumn{1}{c}{\bf Method}  &\multicolumn{1}{c}{\bf LPIPS $\downarrow$} &\multicolumn{1}{c}{\bf DINO $\uparrow$} &\multicolumn{1}{c}{\bf CLIP-T $\uparrow$} &\multicolumn{1}{c}{\bf CLIP-I $\uparrow$}
\\ \hline \\
Textual Inv.~\citep{gal2022image} & 72.35 & 48.92 & 18.76 & 71.23 \\
BLIP~\citep{BLIP-Diffusion}          & 68.21 & 62.96 & \underline{19.68} & 82.38 \\
Omnigen~\citep{xiao2025omnigen}      & 71.68 & 50.05 & 19.41 & 72.95 \\
IP-Adapter~\citep{ye2023ipadapter}  & \underline{62.91} & \underline{72.88} & 19.39 & \underline{89.75} \\
Flux (DiT)~\citep{chang2024fluxfastsoftwarebasedcommunication} & 65.47 & 55.31 & 19.52 & 78.64 \\
AnimalBooth                   & \textbf{49.08} & \textbf{75.66} & \textbf{20.73} & \textbf{90.00} \\
\end{tabular}
\end{center}
\end{table}

\noindent\textbf{Comparison with Optimization-based and DiT Methods.} Table~\ref{tab:quantitative} also compares AnimalBooth with Textual Inversion~\citep{gal2022image} (a representative optimization-based method) and Flux~\citep{chang2024fluxfastsoftwarebasedcommunication} (a state-of-the-art DiT-based model). 
Textual Inversion, despite requiring per-subject optimization, achieves suboptimal results due to limited embedding capacity for complex animal features. Flux, while generating high-quality images, suffers from identity drift when personalizing animal subjects without specialized adaptation. AnimalBooth outperforms both methods across all metrics while requiring no per-subject optimization at inference time.

\begin{figure}[!htbp]
\begin{center}
\includegraphics[width=0.9\linewidth]{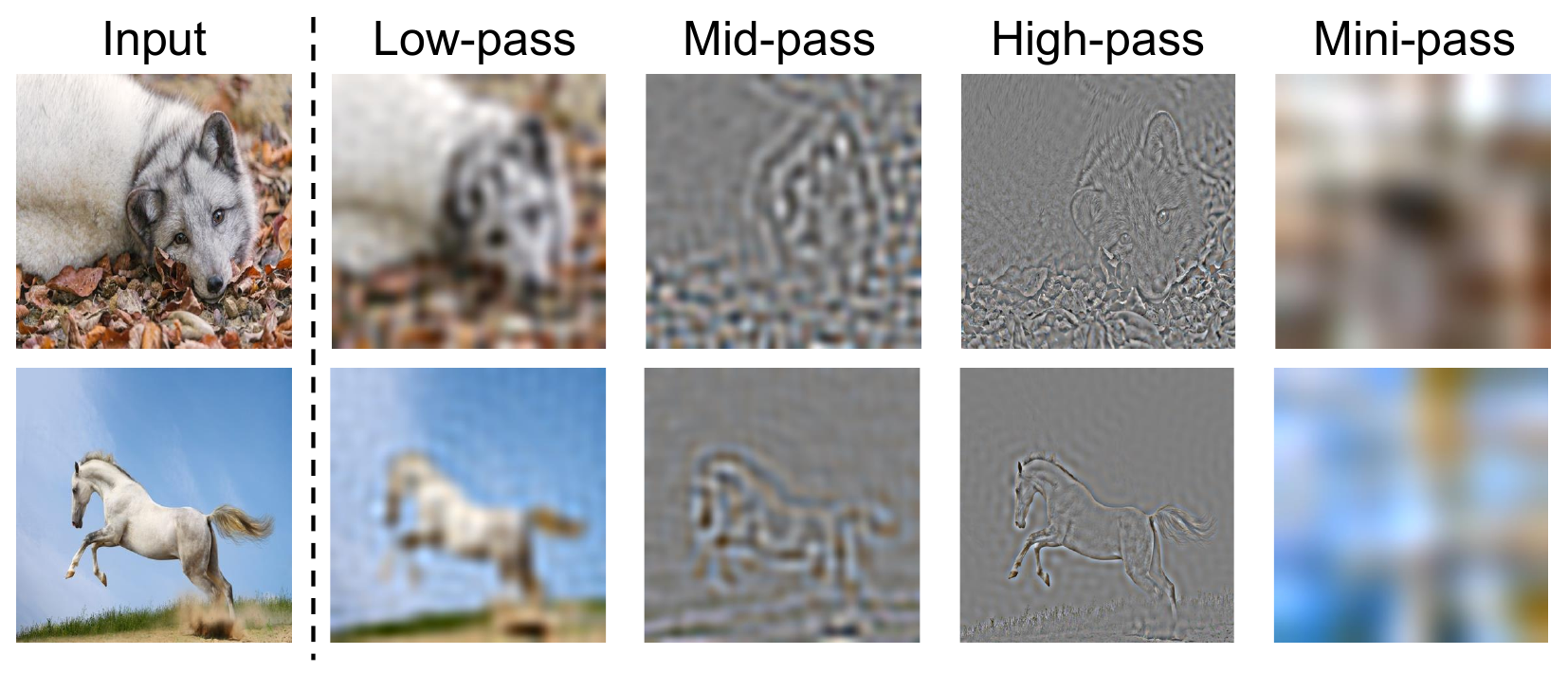}
\end{center}
\vspace{-0.3cm}
  \caption{Output results of different frequency filter masks.}
  \label{fig:visulization}
\end{figure}

\subsection{Ablation Study}
\noindent\textbf{Frequency Configuration.} We evaluate the impact of five distinct frequency configuration modes on generated image quality: Low-Pass, Mini-Pass, Mid-Pass, High-Pass, and without frequency conditioning. As presented in Table~\ref{tab:ablation}, the Low-Pass frequency configuration significantly outperforms other configurations on CLIP-I, DINO and LPIPS metrics, indicating its superior performance in preserving animal identity details such as fur textures, patterns, and coat colors. This is because low-frequency components capture the global structure and morphology essential for identity recognition, while high-frequency details can be effectively reconstructed by the diffusion model. In contrast, while High-Pass shows a slight advantage in semantic structure alignment (CLIP-T), its DINO and CLIP-I scores are notably lower, which is detrimental to the fine-grained restoration of individual animal characteristics.

\noindent\textbf{Projection Layer Design.} Table~\ref{tab:projection} compares different projection layer designs. Q-Former significantly outperforms alternatives, improving CLIP-I by 19.42 points and DINO by 44.61 points. This demonstrates Q-Former's effectiveness as a semantic bottleneck for identity-relevant features.

\noindent\textbf{Guidance Scale.} Table~\ref{tab:guidance} presents the ablation on guidance scale $w$. Low guidance ($w=2.0$) produces poor identity preservation. The optimal $w=7.5$ achieves the best balance between identity fidelity and generation quality.

\noindent\textbf{Hyper-parameter $\lambda$.} Figure~\ref{fig:image_scale} demonstrates the effects of the hyper-parameter $\lambda$ on generated samples with a fixed random seed. As $\lambda$ increases to 1.0, the generated animal gradually loses structural integrity. A smaller $\lambda$ ensures the generated results adhere more closely to the input animal's identity. Consequently, we empirically set $\lambda$ to 0.4 in our experiments.

\noindent\textbf{Visualization.} Fig.~\ref{fig:visulization} demonstrates the impact of different frequency filter masks. The ``Mini-pass'' filter significantly blurs the image, whereas ``Low-pass'' preserves general shapes and smooth color transitions, which is critical for identity preservation. ``Mid-pass'' and ``High-pass'' filters accentuate textural details and sharp edges respectively.

\subsection{Limitations}
Despite these advantages, our method has limitations. The reliance on low-frequency structural guidance can occasionally constrain the diversity of generated poses, particularly for highly dynamic actions that differ significantly from the reference. Additionally, while the Q-Former effectively filters background noise, it may occasionally overlook extremely subtle identity cues that are not semantically salient. Future work will explore extending this framework to animal consistency generation tasks in video modalities.

\section{Conclusion}
\looseness=-1
This paper introduces AnimalBooth, an inference-time tuning-free personalized generation framework specifically designed for animal subjects. Experiments were conducted on our self-constructed AnimalBench dataset, comprising 10,958 training images and 1,000 test images. By integrating the Animal-Net with Q-Former projection and the novel dual-path adaptive attention module, AnimalBooth achieves state-of-the-art performance across all metrics. Furthermore, through the low-pass configuration of the DCT frequency-controlled feature integration module (based on ControlNet), we have enhanced the LPIPS metric by 20 percentage points over baselines. Notably, AnimalBooth achieves 25$\times$ faster inference than DiT-based alternatives while using only 0.7GB VRAM, making it highly practical for real-world deployment. Future work will explore extending this framework to animal consistency generation tasks in video modalities.

\newpage
\bibliography{refs}
\bibliographystyle{iclr2026_delta}
 
\end{document}